\documentclass[11pt,a4paper]{article}
\usepackage[hyperref]{acl2020}
\usepackage{times}
\usepackage{latexsym}
\usepackage{tablefootnote}
\usepackage{graphicx}

\usepackage{amsmath,amsfonts,bm}

\def\eqref#1{equation~\ref{#1}}

\def\1{\bm{1}}

\def\rvp{{\mathbf{p}}}

\def\rvx{{\mathbf{x}}}

\DeclareMathAlphabet{\mathsfit}{\encodingdefault}{\sfdefault}{m}{sl}
\SetMathAlphabet{\mathsfit}{bold}{\encodingdefault}{\sfdefault}{bx}{n}

\newcommand{\E}{\mathbb{E}}

\usepackage{microtype}

\usepackage{booktabs}
\usepackage{amssymb}
\usepackage{amsmath}
\usepackage{enumitem}
\usepackage{mathtools}

\usepackage{hyperref}

\usepackage{subfig}
\usepackage{multirow}
\usepackage{multicol}
\usepackage{xcolor}
\usepackage[hang]{footmisc}
\aclfinalcopy
\def\aclpaperid{1991}

\title{Contextualizing Hate Speech Classifiers with Post-hoc Explanation}

\author{Brendan Kennedy\thanks{Authors contributed equally} \and Xisen Jin\footnotemark[1] \and Aida Mostafazadeh Davani \AND
Morteza Dehghani \and Xiang Ren \\
University of Southern California\\
\texttt {\{btkenned,xisenjin,mostafaz,mdehghan,xiangren\}@usc.edu}}

\begin{document}
\maketitle
\begin{abstract}
Hate speech classifiers trained on imbalanced datasets struggle to determine if group identifiers like ``gay'' or ``black'' are used in offensive or prejudiced ways.
Such biases manifest in false positives when these identifiers are present, due to models' inability to learn the contexts which constitute a hateful usage of identifiers.
We extract SOC~\cite{jin2020towards} post-hoc explanations from fine-tuned BERT classifiers to efficiently detect bias towards identity terms. Then, we propose a novel regularization technique based on these explanations that encourages models to learn from the context of group identifiers in addition to the identifiers themselves.
Our approach improved over baselines in limiting false positives on out-of-domain data while maintaining or improving in-domain performance.\footnotemark[2]\footnotetext[2]{Project page: \url{https://inklab.usc.edu/contextualize-hate-speech/}}
\end{abstract}
\section{Introduction}

Hate speech detection is part of the ongoing effort to limit the harm done by oppressive and abusive language~\citep{waldron2012harm,gelber2016evidencing,gagliardone2015countering,mohan2017impact}. Performance has improved with access to more data and more sophisticated algorithms~\citep[e.g.,][]{mondal2017measurement,silva2016analyzing,del2017hate,basile2019semeval}, but the relative sparsity of hate speech requires sampling using keywords~\citep[e.g.,][]{olteanu2018effect} or sampling from environments with unusually high rates of hate speech~\citep[e.g.,][]{degilbert2018hate,hoover2019bound}. Modern text classifiers thus struggle to learn a model of hate speech that generalizes to real-world applications~\citep{wiegand2019detection}.

\begin{figure}[t]
    \vspace{-0.2cm}
    \centering
    \scalebox{0.98}{
    \includegraphics[width=\linewidth]{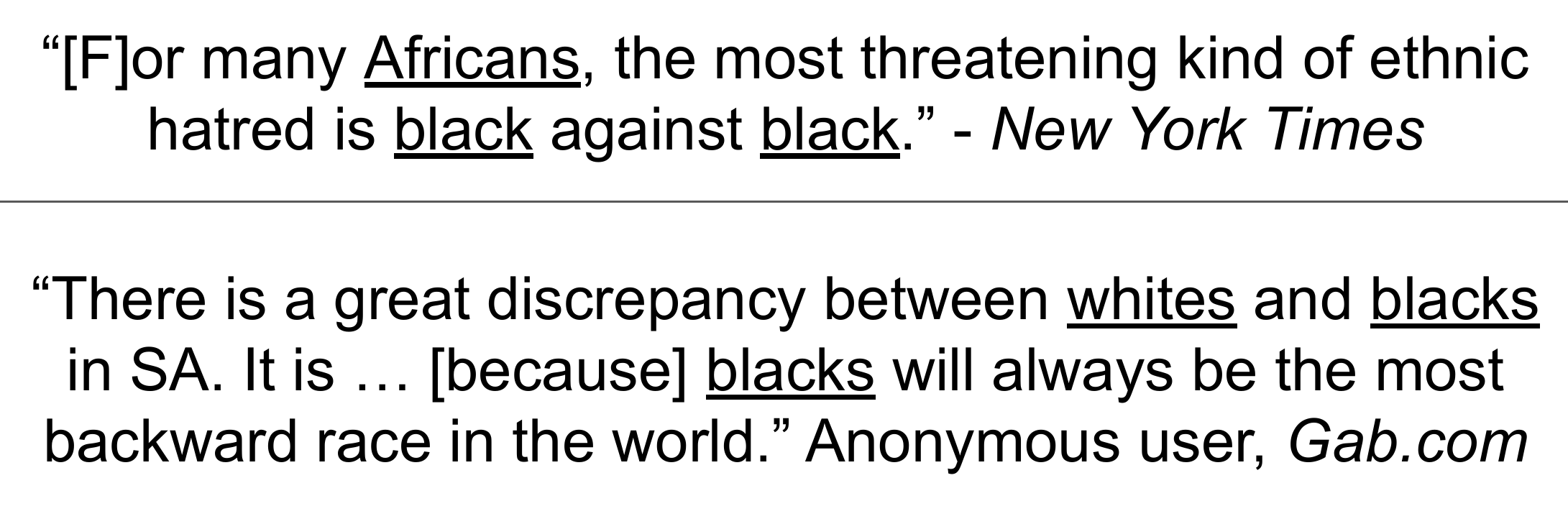}
    }
    \vspace{-0.5cm}
    \caption{Two documents which are classified as hate speech by a fine-tuned BERT classifier. Group identifiers are underlined.}
    \label{fig:examples}
    \vspace{-0.5cm}
\end{figure}

A specific problem found in neural hate speech classifiers is their over-sensitivity to group identifiers like ``Muslim'', ``gay'', and ``black'', which are only hate speech when combined with the right context~\citep{dixon2018measuring}. In Figure \ref{fig:examples} we see two documents containing the word ``black'' that a fine-tuned BERT model predicted to be hate speech, while only the second occurs in a hateful context.

Neural text classifiers achieve state-of-the-art performance in hate speech detection, but are uninterpretable and can break when presented with unexpected inputs~\citep{niven2019probing}. It is thus difficult to contextualize a model's treatment of identifier words. Our approach to this problem is to use the Sampling and Occlusion (SOC) explanation algorithm, which estimates model-agnostic, context-independent post-hoc feature importance~\citep{jin2020towards}. We apply this approach to the Gab Hate Corpus~\citep{kennedy2020gab}, a new corpus labeled for ``hate-based rhetoric'', and an annotated corpus from the Stormfront white supremacist online forum~\citep{degilbert2018hate}.

Based on the explanations generated via SOC, which showed models were biased towards group identifiers, we then propose a novel regularization-based approach in order to increase model sensitivity to the context surrounding group identifiers. We 
regularize importance of group identifiers at training,
coercing models to consider the context surrounding them.

We find that regularization reduces the attention given to group identifiers and heightens the importance of the more generalizable features of hate speech, such as dehumanizing and insulting language. In experiments on an out-of-domain test set of news articles containing group identifiers, which are heuristically assumed to be non-hate speech, we find that regularization greatly reduces the false positive rate, while in-domain, out-of-sample classification performance is either maintained or improved.

\section{Related Work}

Our work is conceptually influenced by \citet{warner2012detecting}, who formulated hate speech detection as disambiguating the use of offensive words from abusive versus non-abusive contexts. More recent approaches applied to a wide typology of hate speech~\citep{waseem2017understanding}, build supervised models trained on annotated~\citep[e.g.,][]{waseem2016hateful,degilbert2018hate} or heuristically-labeled~\citep{wulczyn2017ex,olteanu2018effect} data.
These models suffer from the highly skewed distributions of language in these datasets~\citep{wiegand2019detection}.

Research on bias in classification models also influences this work. \citet{dixon2018measuring} measured and mitigated bias in toxicity classifiers towards social groups, avoiding undesirable predictions of toxicity towards innocuous sentences containing tokens like ``\textit{gay}''. Similarly, annotators' biases towards certain social groups were found to be magnified during classifier training \citet{mostafazadeh2020hatred}.
Specifically within the domain of hate speech and abusive language, \citet{park2018reducing} and \citet{sap2019risk} have defined and studied gender- and racial-bias.
 Techniques for bias reduction in these settings include data augmentation by training on less biased data, term swapping (i.e., swapping gender words), and using debiased word embeddings~\citep{bolukbasi2016man}. 

Complementing these works, we directly manipulate models' modeling of the context surrounding identifier terms by regularizing explanations of these terms. To interpret and modulate fine-tuned language models like BERT, which achieve SotA performance in hate speech detection tasks~\citep{macavaney2019hate,mandl2019overview}, we focus on post-hoc explanation approaches~\citep{guidotti2019survey}. These explanations reveal either word-level~\citep{ribeiro2016should, sundararajan2017axiomatic} or phrase-level importance~\citep{james2018beyond, singh2018hierarchical} of inputs to predictions. ~\cite{rieger2019interpretations, liu2019incorporating} are closely related works in regularizing explanations for fair text classification. However, the explanation methods applied are either incompatible with BERT, or known to be inefficient for regularization as discussed in~\cite{rieger2019interpretations}. We further identify explanations are different in their semantics and compare two explanation algorithms that can be regularized efficiently in our setup. Besides, training by improving counterfactual fairness~\cite{garg2019counterfactual} is another closely related line of works. 

\section{Data}

We selected two public corpora for our experiments which highlight the rhetorical aspects of hate speech, versus merely the usage of slurs and explicitly offensive language~\citep[see][]{davidson2017automated}. The ``Gab Hate Corpus''~\citep[GHC;][]{kennedy2020gab} is a large, random sample (\textit{N} = 27,655) from the Pushshift.io data dump of the Gab network \footnote{\url{https://files.pushshift.io/gab/}}, which we have annotated according to a typology of ``hate-based rhetoric'', a construct motivated by hate speech criminal codes outside the U.S. and social science research on prejudice and dehumanization. Gab is a social network with a high rate of hate speech \cite{zannettou2018gab,lima2018inside} and populated by the ``Alt-right''~\citep{anthony2016inside,benson2016inside}. 
Similarly with respect to domain and definitions, \citet{degilbert2018hate} sampled and annotated posts from the ``Stormfront'' web domain~\citep{meddaugh2009hate} and annotated at the sentence level according to a similar annotation guide as used in the GHC. 

Train and test splits were randomly generated for Stormfront sentences (80/20) with ``hate'' taken as a positive binary label, and a test set was compiled from the GHC by drawing a random stratified sample with respect to the ``target population'' tag (possible values including race/ethnicity target, gender, religious, etc.). A single ``hate'' label was created by taking the union of two main labels, ``human degradation'' and ``calls for violence''. Training data for the GHC (GHC$_{train}$) included 24,353 posts with 2,027 labeled as hate, and test data for the GHC (GHC$_{test}$) included 1,586 posts with 372 labeled as hate. Stormfront splits resulted in 7,896 (1,059 hate) training sentences, 979 (122) validation, and 1,998 (246) test.

\section{Analyzing Group Identifier Bias}

To establish and define our problem more quantitatively, we analyze hate speech models' bias towards group identifiers and how this leads to false positive errors during prediction. We analyze the top features of a linear model and use post-hoc explanations applied to a fine-tuned BERT model in order to measure models' bias towards these terms. We then establish the effect of these tendencies on model predictions using an adversarial-like dataset of New York Times articles. 

\subsection{Classification Models}
We apply our analyses on two text classifiers, logistic regression with bag of words features and a fine-tuned BERT model~\citep{devlin2018bert}. The BERT model appends a special \texttt{CLS} token at the beginning of the input sentence and feeds the sentence into stacked layers of Transformer~\cite{vaswani2017attention} encoders. The representation of the \texttt{CLS} token at the final layer is fed into a linear layer to perform 2-way classification (hate or non-hate).
Model configuration and training details can be found in the Section \ref{appendix:implement_details}.

\subsection{Model Interpretation}\label{internal}

We first determine a model's sensitivity towards group identifiers by examining the models themselves. Linear classifiers can be examined in terms of their most highly-weighted features. We apply a post-hoc explanation algorithm for this task of extracting similar information from the fine-tuned methods discussed above.

\smallskip
\noindent
\textbf{Group identifiers in linear models~~}
From the top features in a bag-of-words logistic regression of hate speech on GHC$_{train}$, we collected a set of twenty-five identity words (not restricted to social group terms, but terms identifying a group in general), including ``homosexual'', ``muslim'', and ``black'', which are used in our later analyses. The full list is in Supplementals (\ref{appendix:term_list}). 
\begin{figure}
    \vspace{-0.2cm}
    \includegraphics[width=\linewidth]{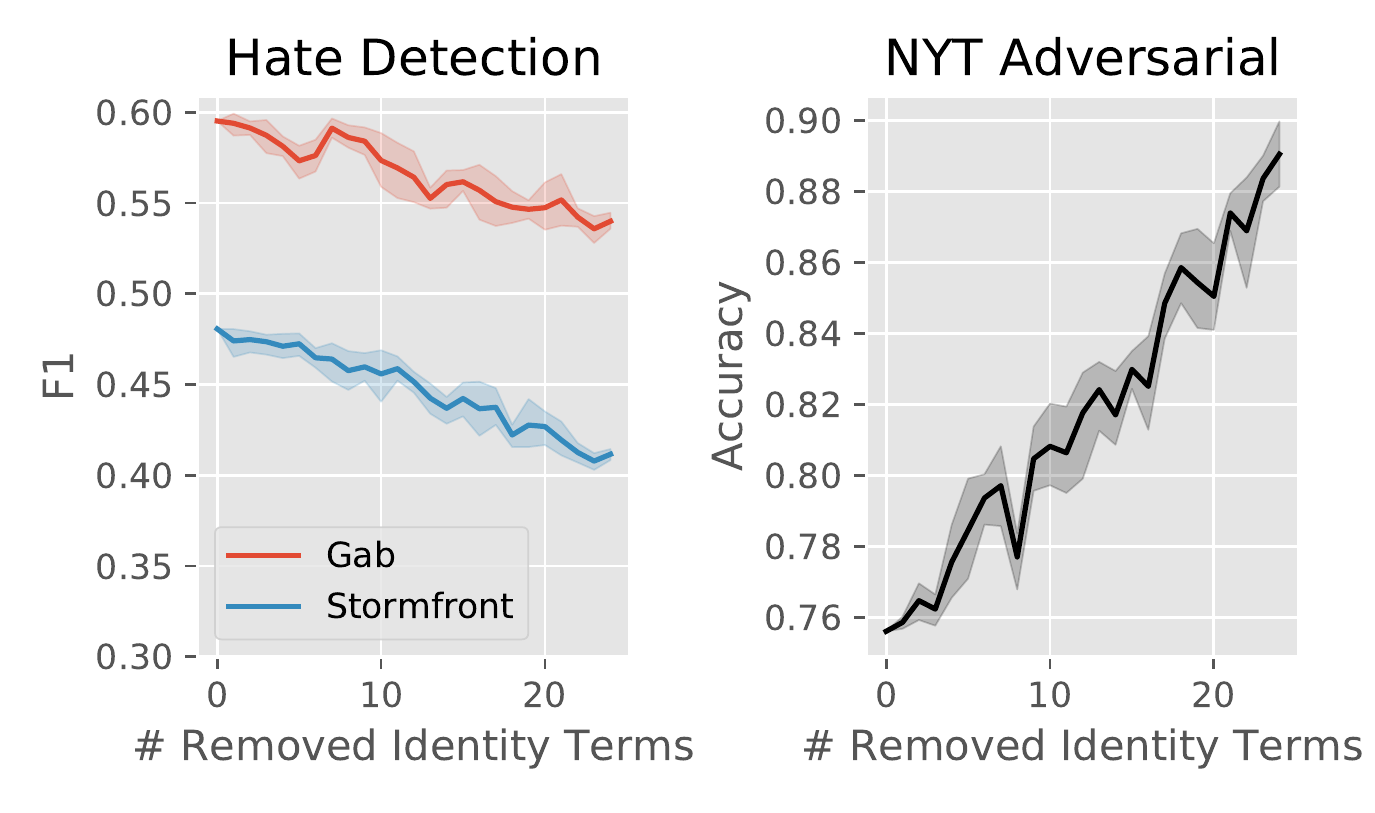}
    \vspace{-0.1cm}
    \caption{BoW F1 scores (trained on GHC$_{train}$ and evaluated on GHC$_{test}$) as a function of how many group identifiers are removed (left). Accuracy of same models on NYT dataset with no hate speech (right).}
    \label{fig:bow_tradeoff}
    \vspace{-0.1cm}
\end{figure}

\smallskip
\noindent
\textbf{Explanation-based measures~~}
\label{sec:explanation_approach}
State-of-the-art fine-tuned BERT models are able to model complicated word and phrase compositions: for example, some words are only offensive when they are composed with specific ethnic groups. 
To capture this, we apply a state-of-the-art Sampling and Occlusion (SOC) algorithm which is capable of generating hierarchical explanations for a prediction.
 
To generate hierarchical explanations, SOC starts by assigning importance score for phrases in a way that eliminates compositional effect between the phrase and its context $\rvx_\delta$ around it within a window. Given a phrase $\rvp$ appearing in a sentence $\rvx$, SOC assigns an importance score $\phi(\rvp)$ to show how the phrase $\rvp$ contribute so that the sentence is classified as a hate speech. The algorithm computes the difference of the unnormalized prediction score $s(\rvx)$ between ``hate'' and ``non-hate'' in the 2-way classifier. Then the algorithm evaluates average change of $s(\rvx)$ when the phrase is masked with padding tokens (noted as $\rvx \backslash \rvp$) for different inputs, in which the $N$-word contexts around the phrase $\rvp$ are sampled from a pretrained language model, while other words remain the same as the given $\rvx$. Formally, the importance score $\phi(\rvp)$ is measured as,
\begin{equation}
\label{sec:contextindependent}
    \phi(\rvp) = \E_{\rvx_\delta}[s(\rvx) - s(\rvx \backslash \rvp)]
\end{equation}
In the meantime, SOC algorithm perform agglomerative clustering over explanations to generate a hierarchical layout.

\smallskip
\noindent
\textbf{Averaged Word-level SOC Explanation~~} 
Using SOC explanations output on GHC$_{test}$, we compute average word importance and present the top 20 in Table \ref{tab:top_reg_words}.


    
    
    

\subsection{Bias in Prediction} 
Hate speech models can be over-attentive to group identifiers, as we have seen by inspecting them through feature analysis and a post-hoc explanation approach.
The effect of this during prediction is that models over-associate these terms with hate speech and choose to neglect the context around the identifier, resulting in false positives.
To provide an external measure of models' over-sensitivity to group identifiers, we construct an adversarial test set of New York Times (NYT) articles that are filtered to contain a balanced, random sample of the twenty-five group identifiers (Section \ref{appendix:term_list}). This gives us $12,500$ documents which are devoid of hate speech as defined by our typologies, excepting quotation.

It is key for models to not \textit{ignore} identifiers, but to match them with the right context.
Figure \ref{fig:bow_tradeoff} shows the effect of ignoring identifiers: random subsets of words ranging in size from $0$ to $25$ are removed, with each subset sample size repeated $5$ times. 
Decreased rates of false positives on the NYT set are accompanied by poor performance in hate speech detection.

\begin{table*}[t]
\vspace{-0.1cm}
    \centering
    \scalebox{0.71}{
    \begin{tabular}{l|cccc|cccc}
         \toprule
         \textbf{Training set} & \multicolumn{4}{c|}{\textbf{GHC}} & \multicolumn{4}{c}{\textbf{Stormfront}} \\ \hline
         \textbf{Method~/~Metrics} & \textbf{Precision} & \textbf{Recall} & \textbf{F1} & \textbf{NYT Acc.} & \textbf{Precision} & \textbf{Recall} & \textbf{F1} & \textbf{NYT Acc.} \\
         \hline
         BoW & 62.80 & 56.72 & 59.60 & 75.61 & 36.95 & 58.13 & 45.18 & 66.78 \\
         BERT & 69.87 $\pm$ 1.7 & 66.83 $\pm$ 7.0 & 67.91 $\pm$ 3.1 & 77.79 $\pm$ 4.8 & \textbf{57.76 $\pm$ 3.9} & 54.43 $\pm$ 8.1 & 55.44 $\pm$ 2.9 & 92.29 $\pm$ 4.1 \\
         \hline
         BoW + WR & 54.65 & 52.15 & 53.37 & 89.72 & 36.24 & 55.69 & 43.91 & 81.34 \\
         BERT + WR & 67.61 $\pm$ 2.8 & 60.08 $\pm$ 6.6 & 63.44 $\pm$ 3.1 & 89.78 $\pm$ 3.8 & 53.16 $\pm$ 4.3  & \textbf{57.03 $\pm$ 5.7}  & 54.60 $\pm$ 1.7 & 92.47 $\pm$ 3.4 \\
         \hline
         BERT + OC ($\alpha$=0.1) & 60.56 $\pm$ 1.8 & \textbf{69.72 $\pm$ 3.6} & 64.14 $\pm$ 3.2  & 89.43 $\pm$ 4.3 & 57.47 $\pm$ 3.7 & 51.10 $\pm$ 4.4 & 53.82 $\pm$ 1.3 & 95.39 $\pm$ 2.3 \\
         BERT + SOC ($\alpha$=0.1) & \textbf{70.17 $\pm$ 2.5}  & 69.03 $\pm$ 3.0 & \textbf{69.52 $\pm$ 1.3} & 83.16 $\pm$ 5.0  & 57.29 $\pm$ 3.4 & 54.27 $\pm$ 3.3 & \textbf{55.55 $\pm$ 1.1} & 93.93 $\pm$ 3.6 \\
         BERT + SOC ($\alpha$=1.0) & 64.29 $\pm$ 3.1 & 69.41 $\pm$ 3.8 & 66.67 $\pm$ 2.5 & \textbf{90.06 $\pm$ 2.6} & 56.05 $\pm$ 3.9 & 54.35 $\pm$ 3.4 & 54.97 $\pm$ 1.1 & \textbf{95.40 $\pm$ 2.0} \\
         \bottomrule
    \end{tabular}
    }
    \caption{Precision, recall, F$_1$ (\%) on GHC$_{test}$ and Stormfront (Stf.) test set and accuracy (\%) on NYT evaluation set. We report mean and standard deviation of the performance across 10 runs for BERT, BERT + WR (word removal), BERT + OC, and BERT + SOC.}
    \label{tab:debias_results}
\end{table*}

\begin{figure}
\vspace{-0.0cm}
    \centering
    \subfloat[][BERT]{\includegraphics[width=0.9\linewidth]{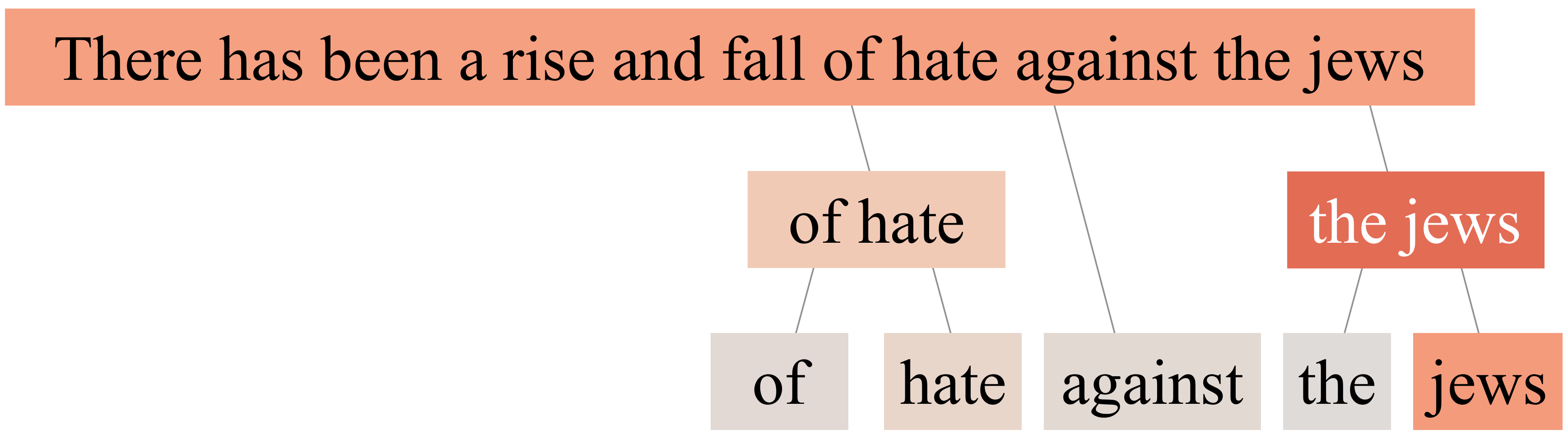}}
    \vspace{-0.3cm}
    \subfloat[][BERT + SOC regularization]{\includegraphics[width=0.9\linewidth]{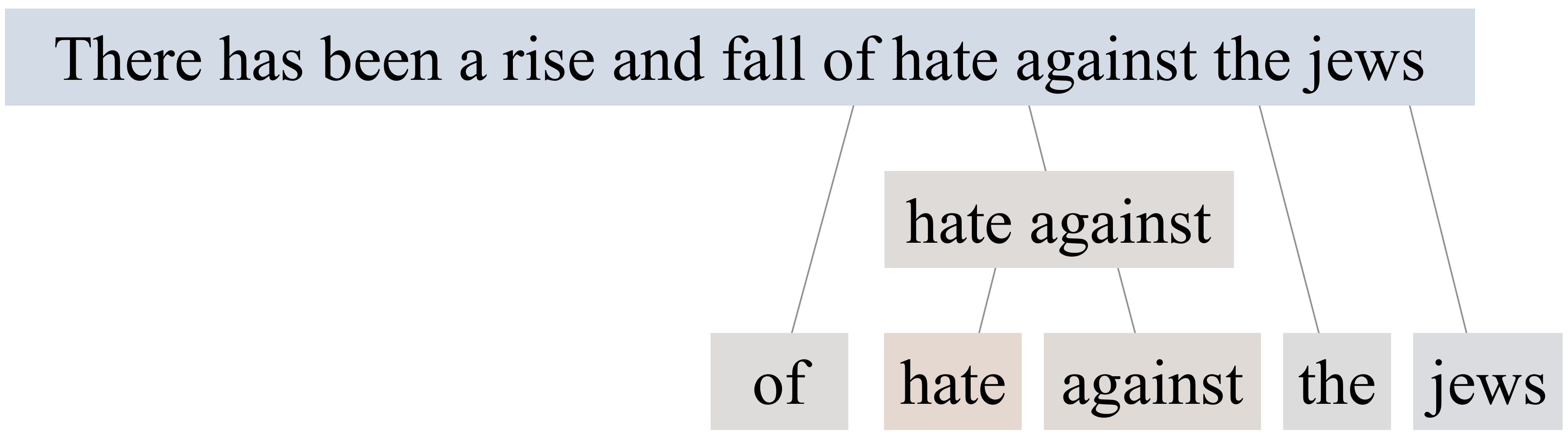}}
    \vspace{-0.1cm}
    \subfloat{\includegraphics[width=0.75\linewidth]{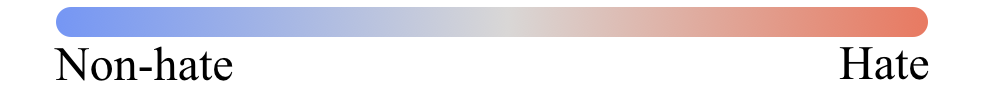}}
    \vspace{-0.1cm}
    \caption{Hierarchical explanations on a test instance from GHC$_{test}$ before and after explanation regularization, where false positive predictions are corrected.}
    \label{fig:vis_reg}
    \vspace{-0.3cm}
\end{figure}

\section{Contextualizing Hate Speech Models}

We have shown hate speech models to be over-sensitive to group identifiers and unable to learn from the context surrounding these words during training. To address this problem in state-of-the-art models, we propose that models can be regularized to give no explained importance to identifier terms. We explain our approach as well as a naive baseline based on removing these terms.

\smallskip
\noindent\textbf{Word Removal Baseline.}
The simplest approach is to remove group identifiers altogether. We remove words from the term list found in Section \ref{appendix:term_list} from 
both training and testing sentences.

\smallskip
\noindent\textbf{Explanation Regularization.}
Given that SOC explanations are fully differentiable, during training, we regularize SOC explanations on the group identifiers to be close to $0$ in addition to the classification objective $\mathcal{L}^\prime$. The combined learning objective is written as follows.
\begin{equation}
    \mathcal{L} = \mathcal{L}^\prime + \alpha\sum_{w \in \rvx \cap S}[\phi(w)]^2,
\end{equation}
where $S$ notes for the set of group names and $\rvx$ notes for the input word sequence. $\alpha$ is a hyperparameter for the strength of the regularization.

In addition to SOC, we also experiment with regularizing input occlusion (OC) explanations, defined as the prediction change when a word or phrase is masked out, which bypass the sampling step in SOC.



\definecolor{lred}{RGB}{255,200,200}
\newcommand{\lred}[1]{\colorbox{lred}{#1}}
\newcommand{\lredit}[1]{\colorbox{lred}{\textit{#1}}}
\begin{table}[t]
\vspace{-0.1cm}
\small
    \centering
    \scalebox{0.9}{
\begin{tabular}{lrlr}
\hline
 \textbf{BERT}  &   \textbf{$\Delta$ Rank} & \textbf{Reg.}       &   \textbf{$\Delta$ Rank} \\
\hline
 ni**er                         &            +0 & ni**er     &            +0 \\
 ni**ers                        &            -7 & fag        &           +35 \\
 kike                           &           -90 & traitor    &           +38 \\
 mosques                        &          -260 & faggot     &            +5 \\
 ni**a                          &          -269 & bastard    &          +814 \\
 \colorbox{lred}{jews}          &          -773 & blamed     &          +294 \\
 kikes                          &          -190 & alive      &         +1013 \\
 nihon                          &          -515 & prostitute &           +56 \\
 faggot                         &            +5 & ni**ers    &            -7 \\
 nip                            &          -314 & undermine  &          +442 \\
 \colorbox{lred}{islam}         &          -882 & punished   &          +491 \\
 \colorbox{lred}{homosexuality} &         -1368 & infection  &         +2556 \\
 nuke                           &          -129 & accusing   &         +2408 \\
 niro                           &          -734 & jaggot     &            +8 \\
 muhammad                       &          -635 & poisoned   &          +357 \\
 faggots                        &          -128 & shitskin   &           +62 \\
 nitrous                        &          -597 & ought      &          +229 \\
 \colorbox{lred}{mexican}       &           -51 & rotting    &          +358 \\
 negro                          &          -346 & stayed     &         +5606 \\
 \colorbox{lred}{muslim}        &         -1855 & destroys   &         +1448 \\
\hline
\end{tabular}
}
\vspace{-0.1cm}
    \caption{Top 20 words by mean SOC weight before (BERT) and after (Reg.) regularization for GHC. Changes in the rank of importance as a result of regularization are also shown. Curated set of group identifiers are highlighted.}
    \label{tab:top_reg_words}
    \vspace{-0.2cm}
\end{table}

\section{Regularization Experiments}
\label{sec:debiasing_exps}
\subsection{Experiment Details}
Balancing performance on hate speech detection and the NYT test set is our quantitative measure of how well a model has learned the contexts in which group identifiers are used for hate speech. We apply our regularization approach to this task, and compare with a word removal strategy for the fine-tuned BERT model. We repeat the process for both the GHC and Stormfront, evaluating test set hate speech classification in-domain and accuracy on the NYT test set. For the GHC, we used the full list of 25 terms; for Stormfront, we used the 10 terms which were also found in the top predictive features in linear classifiers for the Stormfront data. Congruently, for Stormfront we filtered the NYT corpus to only contain these 10 terms (\textit{N} = 5,000).

\subsection{Results}

Performance is reported in Table \ref{tab:debias_results}. 
For the GHC, we see an improvement for in-domain hate speech classification, as well as an improvement in false positive reduction on the NYT corpus. For Stormfront, we see the same improvements for in-domain F$_1$) and NYT. For the GHC, the most marked difference between BERT+WR and BERT+SOC is increased recall, suggesting that baseline removal largely mitigates bias towards identifiers at the cost of more false negatives. 

As discussed in section~\ref{sec:explanation_approach}, SOC eliminates the compositional effects of a given word or phrase. As a result, regularizing SOC explanations does not prohibit the model from utilizing contextual information related to group identifiers. This can possibly explain the improved performance in hate speech detection relative to word removal.

\smallskip
\noindent
\textbf{Word Importance in Regularized Models~~}
We determined that regularization improves a models focus on non-identifier context in prediction. In table \ref{tab:top_reg_words} we show the changes in word importance as measured by SOC. Identity terms' importance decreases, and we also see a significant increase in importance of terms related to hate speech (``poisoned'', ``blamed'', etc.) suggesting that models have learned from the identifier terms' context.

\smallskip
\noindent
\textbf{Visualizing Effects of Regularization~~}
We can further see the effect of regularization by considering Figure \ref{fig:vis_reg}, where hierarchically clustered explanations from SOC are visualized before and after regularization, correcting a false positive.

\section{Conclusion \& Future Work}

Regularizing SOC explanations of group identifiers tunes hate speech classifiers to be more context-sensitive and less reliant on high-frequency words in imbalanced training sets. 
Complementing prior work in bias detection and removal in the context of hate speech and in other settings, our method is directly integrated into Transformer-based models and does not rely on data augmentation. As such, it is an encouraging technique towards directing models' internal representation of target phenomena via lexical anchors.

Future work includes direct extension and validation of this technique with other language models such as GPT-2~\citep{radford2019language}; experimenting with other hate speech or offensive language datasets; and experimenting with these and other sets of identity terms. Also motivated by the present work is the more general pursuit of integrating structure into neural models like BERT.

Regularized hate speech classifiers increases sensitivity to the compositionality of hate speech, but the phenomena remain highly complex rhetorically and difficult to learn through supervision. For example, this post from the GHC requires background information and reasoning across sentences in order to classify as offensive or prejudiced: ``Donald Trump received much criticism for referring to Haiti, El Salvador and Africa as `shitholes'. He was simply speaking the truth.'' The examples we presented (see Appendix \ref{fig:vis_appendix_1} and \ref{fig:vis_appendix_2}) show that regularization leads to models that are context-sensitive to a degree, but not to the extent of reasoning over sentences like those above. We hope that the present work can motivate more attempts to inject more structure into hate speech classification.

Explanation algorithms offer a window into complex predictive models, and regularization as performed in this work can improve models' internal representations of target phenomena. 

\section*{Acknowledgments}
This research was sponsored in part by NSF CAREER BCS-1846531 (Morteza Dehghani).
Xiang Ren's research is based upon work supported in part by the Office of the Director of National Intelligence (ODNI), Intelligence Advanced Research Projects Activity (IARPA), via Contract No. 2019-19051600007, United States Office Of Naval Research under Contract No. N660011924033, and NSF SMA 18-29268.

\bibliography{acl2020}

\begin{thebibliography}{44}
\expandafter\ifx\csname natexlab\endcsname\relax\def\natexlab#1{#1}\fi

\bibitem[{Anthony(2016)}]{anthony2016inside}
Andrew Anthony. 2016.
\newblock Inside the hate-filled echo chamber of racism and conspiracy
  theories.
\newblock \emph{The guardian}, 18.

\bibitem[{Basile et~al.(2019)Basile, Bosco, Fersini, Nozza, Patti, Pardo,
  Rosso, and Sanguinetti}]{basile2019semeval}
Valerio Basile, Cristina Bosco, Elisabetta Fersini, Debora Nozza, Viviana
  Patti, Francisco Manuel~Rangel Pardo, Paolo Rosso, and Manuela Sanguinetti.
  2019.
\newblock Semeval-2019 task 5: Multilingual detection of hate speech against
  immigrants and women in twitter.
\newblock In \emph{Proceedings of the 13th International Workshop on Semantic
  Evaluation}, pages 54--63.

\bibitem[{Benson(2016)}]{benson2016inside}
Thor Benson. 2016.
\newblock Inside the twitter for racists: Gab the site where milo yiannopoulos
  goes to troll now.

\bibitem[{Bolukbasi et~al.(2016)Bolukbasi, Chang, Zou, Saligrama, and
  Kalai}]{bolukbasi2016man}
Tolga Bolukbasi, Kai-Wei Chang, James~Y Zou, Venkatesh Saligrama, and Adam~T
  Kalai. 2016.
\newblock Man is to computer programmer as woman is to homemaker? debiasing
  word embeddings.
\newblock In \emph{Advances in neural information processing systems}, pages
  4349--4357.

\bibitem[{Davidson et~al.(2017)Davidson, Warmsley, Macy, and
  Weber}]{davidson2017automated}
Thomas Davidson, Dana Warmsley, Michael Macy, and Ingmar Weber. 2017.
\newblock Automated hate speech detection and the problem of offensive
  language.
\newblock In \emph{Eleventh international {AAAI} conference on web and social
  media}.

\bibitem[{Del~Vigna12 et~al.(2017)Del~Vigna12, Cimino23, Dell’Orletta,
  Petrocchi, and Tesconi}]{del2017hate}
Fabio Del~Vigna12, Andrea Cimino23, Felice Dell’Orletta, Marinella Petrocchi,
  and Maurizio Tesconi. 2017.
\newblock Hate me, hate me not: Hate speech detection on facebook.
\newblock In \emph{Proceedings of the First Italian Conference on Cybersecurity
  (ITASEC17)}, pages 86--95.

\bibitem[{Devlin et~al.(2018)Devlin, Chang, Lee, and
  Toutanova}]{devlin2018bert}
Jacob Devlin, Ming-Wei Chang, Kenton Lee, and Kristina Toutanova. 2018.
\newblock Bert: Pre-training of deep bidirectional transformers for language
  understanding.
\newblock \emph{arXiv preprint arXiv:1810.04805}.

\bibitem[{Dixon et~al.(2018)Dixon, Li, Sorensen, Thain, and
  Vasserman}]{dixon2018measuring}
Lucas Dixon, John Li, Jeffrey Sorensen, Nithum Thain, and Lucy Vasserman. 2018.
\newblock Measuring and mitigating unintended bias in text classification.
\newblock In \emph{Proceedings of the 2018 AAAI/ACM Conference on AI, Ethics,
  and Society}, pages 67--73. ACM.

\bibitem[{Gagliardone et~al.(2015)Gagliardone, Gal, Alves, and
  Martinez}]{gagliardone2015countering}
Iginio Gagliardone, Danit Gal, Thiago Alves, and Gabriela Martinez. 2015.
\newblock \emph{Countering online hate speech}.
\newblock Unesco Publishing.

\bibitem[{Garg et~al.(2019)Garg, Perot, Limtiaco, Taly, Chi, and
  Beutel}]{garg2019counterfactual}
Sahaj Garg, Vincent Perot, Nicole Limtiaco, Ankur Taly, Ed~H Chi, and Alex
  Beutel. 2019.
\newblock Counterfactual fairness in text classification through robustness.
\newblock In \emph{Proceedings of the 2019 AAAI/ACM Conference on AI, Ethics,
  and Society}, pages 219--226.

\bibitem[{Gelber and McNamara(2016)}]{gelber2016evidencing}
Katharine Gelber and Luke McNamara. 2016.
\newblock Evidencing the harms of hate speech.
\newblock \emph{Social Identities}, 22(3):324--341.

\bibitem[{de~Gibert et~al.(2018)de~Gibert, Perez, Pablos, and
  Cuadros}]{degilbert2018hate}
Ona de~Gibert, Naiara Perez, Aitor~Garc{\'\i}a Pablos, and Montse Cuadros.
  2018.
\newblock Hate speech dataset from a white supremacy forum.
\newblock In \emph{Proceedings of the 2nd Workshop on Abusive Language Online
  (ALW2)}, pages 11--20.

\bibitem[{Guidotti et~al.(2019)Guidotti, Monreale, Ruggieri, Turini, Giannotti,
  and Pedreschi}]{guidotti2019survey}
Riccardo Guidotti, Anna Monreale, Salvatore Ruggieri, Franco Turini, Fosca
  Giannotti, and Dino Pedreschi. 2019.
\newblock A survey of methods for explaining black box models.
\newblock \emph{ACM computing surveys (CSUR)}, 51(5):93.

\bibitem[{Hoover et~al.(2019)Hoover, Atari, Davani, Kennedy, Portillo-Wightman,
  Yeh, Kogon, and Dehghani}]{hoover2019bound}
Joseph Hoover, Mohammad Atari, Aida~Mostafazadeh Davani, Brendan Kennedy,
  Gwenyth Portillo-Wightman, Leigh Yeh, Drew Kogon, and Morteza Dehghani. 2019.
\newblock Bound in hatred: The role of group-based morality in acts of hate.
\newblock \emph{PsyArxiv Preprint 10.31234/osf.io/359me}.

\bibitem[{Jin et~al.(2020)Jin, Wei, Du, Xue, and Ren}]{jin2020towards}
Xisen Jin, Zhongyu Wei, Junyi Du, Xiangyang Xue, and Xiang Ren. 2020.
\newblock \href {https://openreview.net/forum?id=BkxRRkSKwr} {Towards
  hierarchical importance attribution: Explaining compositional semantics for
  neural sequence models}.
\newblock In \emph{International Conference on Learning Representations}.

\bibitem[{Kennedy et~al.(2020)Kennedy, Atari, Davani, Yeh, Omrani, Kim,
  Coombs~Jr., Havaldar, Portillo-Wightman, Gonzalez, Hoover, Azatian, Cardenas,
  Hussain, Lara, Omary, Park, Wang, Wijaya, Zhang, Meyerowitz, and
  Dehghani}]{kennedy2020gab}
Brendan Kennedy, Mohammad Atari, Aida~M Davani, Leigh Yeh, Ali Omrani, Yehsong
  Kim, Kris Coombs~Jr., Shreya Havaldar, Gwenyth Portillo-Wightman, Elaine
  Gonzalez, Joe Hoover, Aida Azatian, Gabriel Cardenas, Alyzeh Hussain, Austin
  Lara, Adam Omary, Christina Park, Xin Wang, Clarisa Wijaya, Yong Zhang, Beth
  Meyerowitz, and Morteza Dehghani. 2020.
\newblock \href {https://doi.org/10.31234/osf.io/hqjxn} {The gab hate corpus: A
  collection of 27k posts annotated for hate speech}.

\bibitem[{Kingma and Ba(2015)}]{kingma2014adam}
Diederik~P Kingma and Jimmy Ba. 2015.
\newblock Adam: A method for stochastic optimization.
\newblock In \emph{International Conference on Learning Representations}.

\bibitem[{Lima et~al.(2018)Lima, Reis, Melo, Murai, Araujo, Vikatos, and
  Benevenuto}]{lima2018inside}
Lucas Lima, Julio~CS Reis, Philipe Melo, Fabricio Murai, Leandro Araujo,
  Pantelis Vikatos, and Fabricio Benevenuto. 2018.
\newblock Inside the right-leaning echo chambers: Characterizing gab, an
  unmoderated social system.
\newblock In \emph{2018 IEEE/ACM International Conference on Advances in Social
  Networks Analysis and Mining (ASONAM)}, pages 515--522. IEEE.

\bibitem[{Liu and Avci(2019)}]{liu2019incorporating}
Frederick Liu and Besim Avci. 2019.
\newblock Incorporating priors with feature attribution on text classification.
\newblock \emph{arXiv preprint arXiv:1906.08286}.

\bibitem[{MacAvaney et~al.(2019)MacAvaney, Yao, Yang, Russell, Goharian, and
  Frieder}]{macavaney2019hate}
Sean MacAvaney, Hao-Ren Yao, Eugene Yang, Katina Russell, Nazli Goharian, and
  Ophir Frieder. 2019.
\newblock Hate speech detection: Challenges and solutions.
\newblock \emph{PloS one}, 14(8).

\bibitem[{Mandl et~al.(2019)Mandl, Modha, Majumder, Patel, Dave, Mandlia, and
  Patel}]{mandl2019overview}
Thomas Mandl, Sandip Modha, Prasenjit Majumder, Daksh Patel, Mohana Dave,
  Chintak Mandlia, and Aditya Patel. 2019.
\newblock Overview of the hasoc track at fire 2019: Hate speech and offensive
  content identification in indo-european languages.
\newblock In \emph{Proceedings of the 11th Forum for Information Retrieval
  Evaluation}, pages 14--17.

\bibitem[{Meddaugh and Kay(2009)}]{meddaugh2009hate}
Priscilla~Marie Meddaugh and Jack Kay. 2009.
\newblock Hate speech or ``reasonable racism?'' the other in stormfront.
\newblock \emph{Journal of Mass Media Ethics}, 24(4):251--268.

\bibitem[{Mohan et~al.(2017)Mohan, Guha, Harris, Popowich, Schuster, and
  Priebe}]{mohan2017impact}
Shruthi Mohan, Apala Guha, Michael Harris, Fred Popowich, Ashley Schuster, and
  Chris Priebe. 2017.
\newblock The impact of toxic language on the health of reddit communities.
\newblock In \emph{Canadian Conference on Artificial Intelligence}, pages
  51--56. Springer.

\bibitem[{Mondal et~al.(2017)Mondal, Silva, and
  Benevenuto}]{mondal2017measurement}
Mainack Mondal, Leandro~Ara{\'u}jo Silva, and Fabr{\'\i}cio Benevenuto. 2017.
\newblock A measurement study of hate speech in social media.
\newblock In \emph{Proceedings of the 28th ACM Conference on Hypertext and
  Social Media}, pages 85--94. ACM.

\bibitem[{Mostafazadeh~Davani et~al.(2020)Mostafazadeh~Davani, Atari, Kennedy,
  Havaldar, and Dehghani}]{mostafazadeh2020hatred}
Aida Mostafazadeh~Davani, Mohammad Atari, Brendan Kennedy, Shreya Havaldar, and
  Morteza Dehghani. 2020.
\newblock Hatred is in the eye of the annotator: Hate speech classifiers learn
  human-like social stereotypes (in press).
\newblock In \emph{31st Annual Conference of the Cognitive Science Society
  (CogSci)}.

\bibitem[{Murdoch et~al.(2018)Murdoch, Liu, and Yu}]{james2018beyond}
W.~James Murdoch, Peter~J. Liu, and Bin Yu. 2018.
\newblock \href {https://openreview.net/forum?id=rkRwGg-0Z} {Beyond word
  importance: Contextual decomposition to extract interactions from {LSTM}s}.
\newblock In \emph{International Conference on Learning Representations}.

\bibitem[{Niven and Kao(2019)}]{niven2019probing}
Timothy Niven and Hung-Yu Kao. 2019.
\newblock Probing neural network comprehension of natural language arguments.
\newblock In \emph{Proceedings of the 57th Annual Meeting of the Association
  for Computational Linguistics}, pages 4658--4664.

\bibitem[{Olteanu et~al.(2018)Olteanu, Castillo, Boy, and
  Varshney}]{olteanu2018effect}
Alexandra Olteanu, Carlos Castillo, Jeremy Boy, and Kush~R Varshney. 2018.
\newblock The effect of extremist violence on hateful speech online.
\newblock In \emph{Twelfth International AAAI Conference on Web and Social
  Media}.

\bibitem[{Park et~al.(2018)Park, Shin, and Fung}]{park2018reducing}
Ji~Ho Park, Jamin Shin, and Pascale Fung. 2018.
\newblock Reducing gender bias in abusive language detection.
\newblock In \emph{Proceedings of the 2018 Conference on Empirical Methods in
  Natural Language Processing}, pages 2799--2804.

\bibitem[{Radford et~al.(2019)Radford, Wu, Child, Luan, Amodei, and
  Sutskever}]{radford2019language}
Alec Radford, Jeffrey Wu, Rewon Child, David Luan, Dario Amodei, and Ilya
  Sutskever. 2019.
\newblock Language models are unsupervised multitask learners.
\newblock \emph{Open {AI} Blog}.

\bibitem[{Ribeiro et~al.(2016)Ribeiro, Singh, and Guestrin}]{ribeiro2016should}
Marco~Tulio Ribeiro, Sameer Singh, and Carlos Guestrin. 2016.
\newblock Why should i trust you?: Explaining the predictions of any
  classifier.
\newblock In \emph{Proceedings of the 22nd ACM SIGKDD international conference
  on knowledge discovery and data mining}, pages 1135--1144. ACM.

\bibitem[{Rieger et~al.(2019)Rieger, Singh, Murdoch, and
  Yu}]{rieger2019interpretations}
Laura Rieger, Chandan Singh, W~James Murdoch, and Bin Yu. 2019.
\newblock Interpretations are useful: penalizing explanations to align neural
  networks with prior knowledge.
\newblock \emph{arXiv preprint arXiv:1909.13584}.

\bibitem[{Sap et~al.(2019)Sap, Card, Gabriel, Choi, and Smith}]{sap2019risk}
Maarten Sap, Dallas Card, Saadia Gabriel, Yejin Choi, and Noah~A Smith. 2019.
\newblock The risk of racial bias in hate speech detection.
\newblock In \emph{Proceedings of the 57th Annual Meeting of the Association
  for Computational Linguistics}, pages 1668--1678.

\bibitem[{Silva et~al.(2016)Silva, Mondal, Correa, Benevenuto, and
  Weber}]{silva2016analyzing}
Leandro Silva, Mainack Mondal, Denzil Correa, Fabr{\'\i}cio Benevenuto, and
  Ingmar Weber. 2016.
\newblock Analyzing the targets of hate in online social media.
\newblock In \emph{Tenth International AAAI Conference on Web and Social
  Media}.

\bibitem[{Singh et~al.(2019)Singh, Murdoch, and Yu}]{singh2018hierarchical}
Chandan Singh, W.~James Murdoch, and Bin Yu. 2019.
\newblock \href {https://openreview.net/forum?id=SkEqro0ctQ} {Hierarchical
  interpretations for neural network predictions}.
\newblock In \emph{International Conference on Learning Representations}.

\bibitem[{Sundararajan et~al.(2017)Sundararajan, Taly, and
  Yan}]{sundararajan2017axiomatic}
Mukund Sundararajan, Ankur Taly, and Qiqi Yan. 2017.
\newblock Axiomatic attribution for deep networks.
\newblock In \emph{Proceedings of the 34th International Conference on Machine
  Learning-Volume 70}, pages 3319--3328. JMLR. org.

\bibitem[{Vaswani et~al.(2017)Vaswani, Shazeer, Parmar, Uszkoreit, Jones,
  Gomez, Kaiser, and Polosukhin}]{vaswani2017attention}
Ashish Vaswani, Noam Shazeer, Niki Parmar, Jakob Uszkoreit, Llion Jones,
  Aidan~N Gomez, {\L}ukasz Kaiser, and Illia Polosukhin. 2017.
\newblock Attention is all you need.
\newblock In \emph{Advances in neural information processing systems}, pages
  5998--6008.

\bibitem[{Waldron(2012)}]{waldron2012harm}
Jeremy Waldron. 2012.
\newblock \emph{The harm in hate speech}.
\newblock Harvard University Press.

\bibitem[{Warner and Hirschberg(2012)}]{warner2012detecting}
William Warner and Julia Hirschberg. 2012.
\newblock Detecting hate speech on the world wide web.
\newblock In \emph{Proceedings of the second workshop on language in social
  media}, pages 19--26. Association for Computational Linguistics.

\bibitem[{Waseem et~al.(2017)Waseem, Davidson, Warmsley, and
  Weber}]{waseem2017understanding}
Zeerak Waseem, Thomas Davidson, Dana Warmsley, and Ingmar Weber. 2017.
\newblock Understanding abuse: A typology of abusive language detection
  subtasks.
\newblock In \emph{Proceedings of the First Workshop on Abusive Language
  Online}, pages 78--84.

\bibitem[{Waseem and Hovy(2016)}]{waseem2016hateful}
Zeerak Waseem and Dirk Hovy. 2016.
\newblock Hateful symbols or hateful people? predictive features for hate
  speech detection on twitter.
\newblock In \emph{Proceedings of the NAACL student research workshop}, pages
  88--93.

\bibitem[{Wiegand et~al.(2019)Wiegand, Ruppenhofer, and
  Kleinbauer}]{wiegand2019detection}
Michael Wiegand, Josef Ruppenhofer, and Thomas Kleinbauer. 2019.
\newblock Detection of abusive language: the problem of biased datasets.
\newblock In \emph{Proceedings of the 2019 Conference of the North American
  Chapter of the Association for Computational Linguistics: Human Language
  Technologies, Volume 1 (Long and Short Papers)}, pages 602--608.

\bibitem[{Wulczyn et~al.(2017)Wulczyn, Thain, and Dixon}]{wulczyn2017ex}
Ellery Wulczyn, Nithum Thain, and Lucas Dixon. 2017.
\newblock Ex machina: Personal attacks seen at scale.
\newblock In \emph{Proceedings of the 26th International Conference on World
  Wide Web}, pages 1391--1399. International World Wide Web Conferences
  Steering Committee.

\bibitem[{Zannettou et~al.(2018)Zannettou, Bradlyn, De~Cristofaro, Kwak,
  Sirivianos, Stringini, and Blackburn}]{zannettou2018gab}
Savvas Zannettou, Barry Bradlyn, Emiliano De~Cristofaro, Haewoon Kwak, Michael
  Sirivianos, Gianluca Stringini, and Jeremy Blackburn. 2018.
\newblock What is gab: A bastion of free speech or an alt-right echo chamber.
\newblock In \emph{Companion Proceedings of the The Web Conference 2018}, pages
  1007--1014. International World Wide Web Conferences Steering Committee.

\end{thebibliography}
\bibliographystyle{acl_natbib}

\clearpage
\appendix

\section{Appendices}
\label{sec:appendix}

\subsection{Full List of Curated Group Identifiers} \label{appendix:term_list}

\begin{table}[!ht]
    \centering
    \scalebox{0.9}{
    \begin{tabular}{p{\linewidth}}
    \hline
     muslim jew jews white islam blacks muslims women whites gay black democat islamic allah jewish lesbian transgender race brown woman mexican religion homosexual homosexuality africans \\ \hline
    \end{tabular}
    }
    \caption{25 group identifiers selected from top weighted words in the TF-IDF BOW linear classifier on the GHC.}
    \label{tab:curated_word_list_ghc}
\end{table}

\begin{table}[!ht]
    \centering
    \scalebox{0.9}{
    \begin{tabular}{p{\linewidth}}
    \hline
     jew jews mexican blacks jewish brown black muslim homosexual islam \\ \hline
    \end{tabular}
    }
    \caption{10 group identifiers selected for the Stormfront dataset.}
    \label{tab:curated_word_list_stormfront}
\end{table}

\subsection{Visualizations of Effect of Regularization}
\begin{figure}[!ht]
    \centering
    \subfloat[][BERT]{\includegraphics[width=\linewidth]{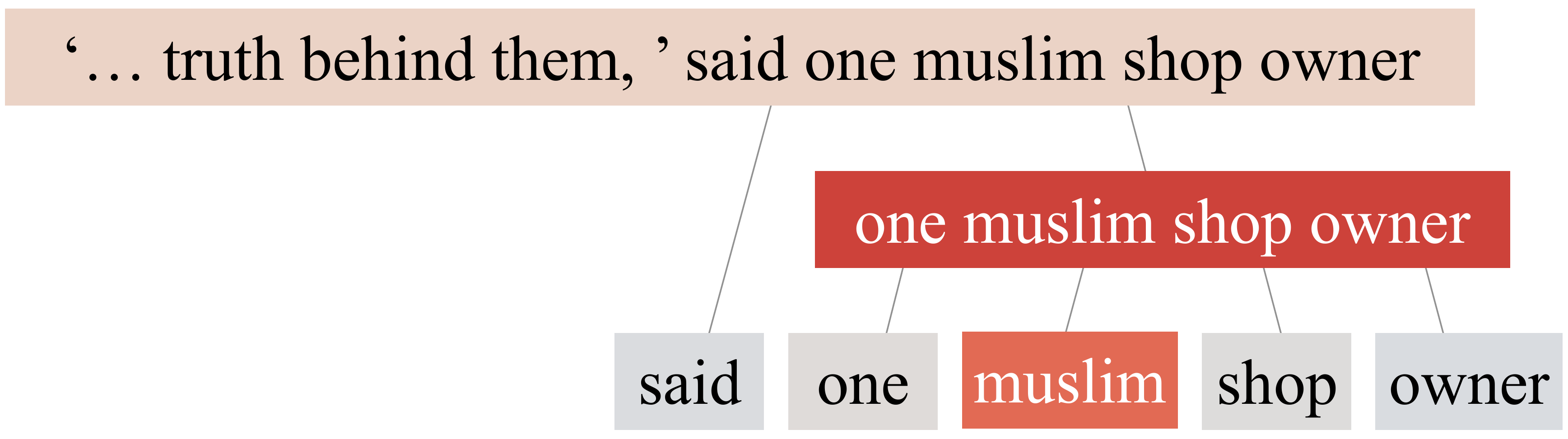}}
    \vspace{-0.3cm}
    \subfloat[][BERT + SOC regularization]{\includegraphics[width=\linewidth]{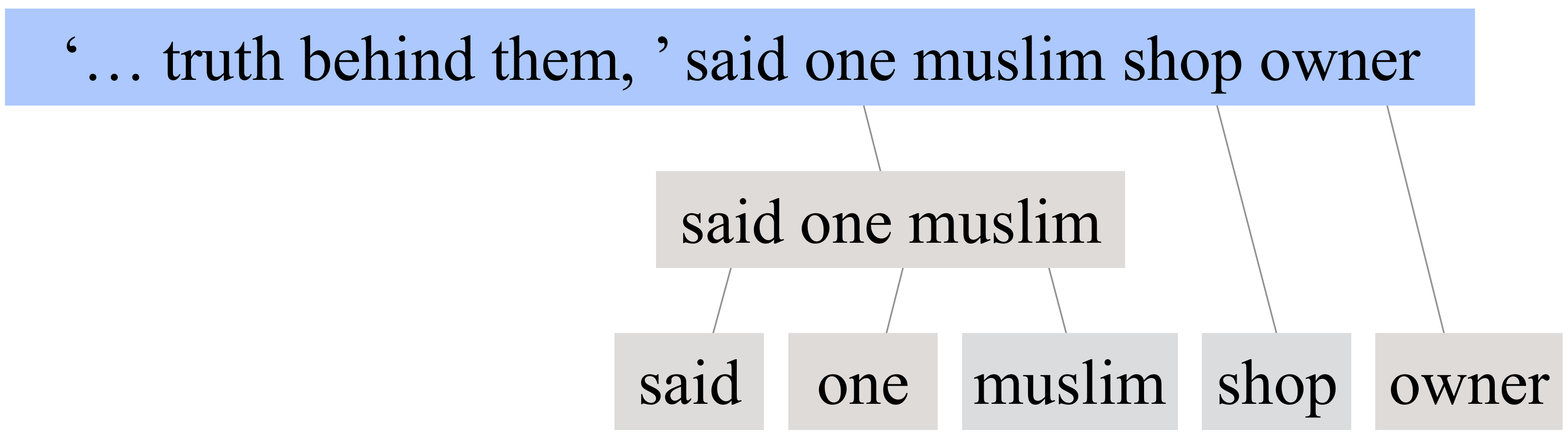}}
    \vspace{-0.1cm}
    \subfloat{\includegraphics[width=0.75\linewidth]{figures/soc_coolwarm/bar.png}}
    \vspace{-0.2cm}
    \caption{Hierarchical explanations on a test instance from the NYT dataset where false positive predictions are corrected.}
    \label{fig:vis_appendix_1}
\end{figure}

\begin{figure}[!ht]
    \centering
    \subfloat[][BERT]{\includegraphics[width=\linewidth]{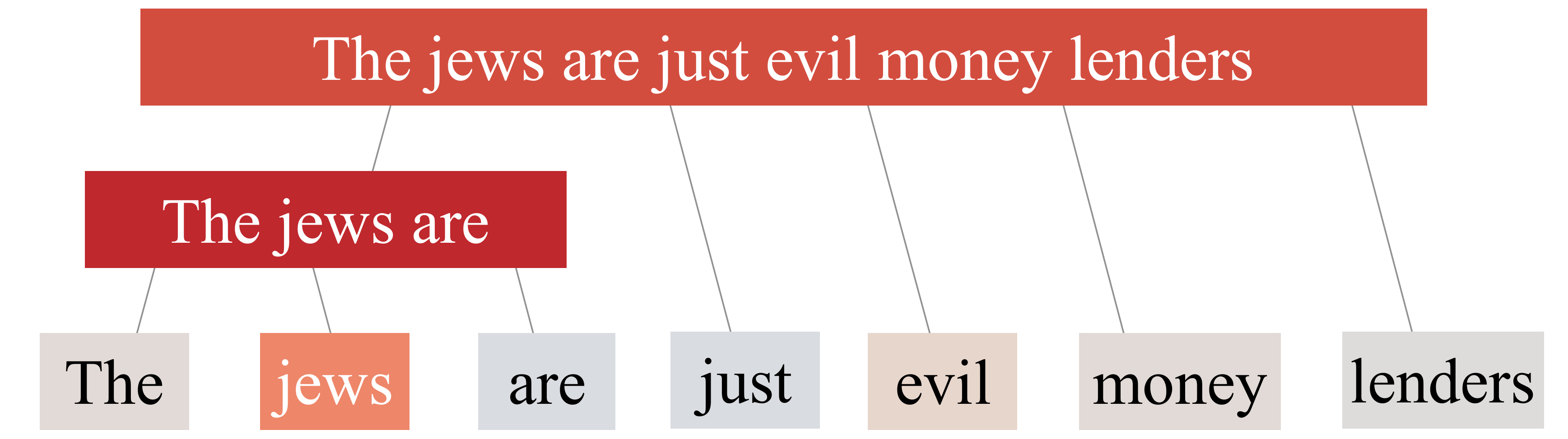}}
    \vspace{-0.3cm}
    \subfloat[][BERT + SOC regularization]{\includegraphics[width=\linewidth]{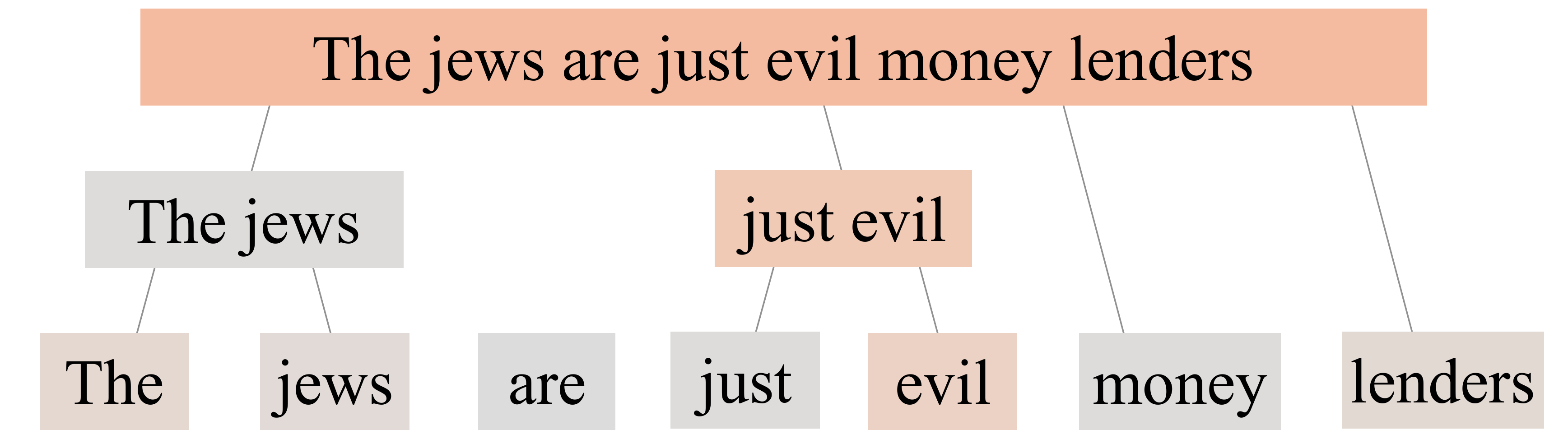}}
    \vspace{-0.1cm}
    \subfloat{\includegraphics[width=0.75\linewidth]{figures/soc_coolwarm/bar.png}}
    \vspace{-0.2cm}
    \caption{Hierarchical explanations on a test instance from the Gab dataset where both models make correct positive predictions. However, the explanations reveal that only the regularized model is making correct predictions for correct reasons.}
    \label{fig:vis_appendix_2}
\end{figure}

\subsection{Implementation Details} \label{appendix:implement_details}
\noindent \textbf{Training Details.} We fine-tune over the BERT-base model using the public code\footnote{\url{https://github.com/huggingface/transformers}}, where the batch size is set to 32 and the learning rate of the Adam~\citep{kingma2014adam} optimizer is set to $2 \times 10^{-5}$. The validation is performed every 200 iterations and the learning rate is halved when the validation F1 decreases. The training stops when the learning rate is halved for 5 times. To handle the data imbalance issue, we reweight the training loss so that positive examples are weighted 10 times as negative examples on the Gab dataset and 8 times on the Stormfront dataset.

\smallskip
\noindent \textbf{Explanation Algorithm Details.} For the SOC algorithm, we set the number of samples and the size of the context window as 20 and 20 respectively for explanation analysis, and set two parameters as 5 and 5 respectively for explanation regularization.

\subsection{Cross-Domain Performance}

In addition to evaluating each model within-domain (i.e., training on GHC$_{train}$ and evaluating on GHC$_{test}$) we evaluated each model across domains. The results of these experiments, conducted in the same way as before, are presented in Table~\ref{tab:debias_results_appendix}.

\begin{table}[!ht]
\vspace{-0.1cm}
    \centering
    \scalebox{0.80}{
    \begin{tabular}{lcc}
         \toprule
         \textbf{Method~/~Dataset} & \textbf{Gab $\rightarrow$ Stf. F1} & \textbf{Stf. $\rightarrow$ Gab F1} \\
         \midrule
         BoW & 32.39 & 46.71 \\
         BERT & 42.84 $\pm$ 1.2 & 53.80 $\pm$ 5.5 \\
         \midrule
         BoW + WR & 27.45  & 44.81 \\
         BERT + WR & 39.10 $\pm$ 1.3 & 55.31 $\pm$ 4.0 \\
         \midrule 
         BERT + OC ($\alpha$=0.1) & 40.60 $\pm$ 1.6 & 56.90 $\pm$ 1.8  \\
         BERT + SOC ($\alpha$=0.1) & 41.88 $\pm$ 1.0 & 55.75 $\pm$ 2.1  \\
         BERT + SOC ($\alpha$=1.0) & 39.20 $\pm$ 2.7 & 56.82 $\pm$ 3.9 \\
         \bottomrule
    \end{tabular}
    }
    \vspace{-0.2cm}
    \caption{Cross domain F1 on Gab, Stormfront (Stf.) datasets. We report mean and standard deviation of the performance within 10 runs for BERT, BERT + WR (word removal), BERT + OC, and BERT + SOC.}
    \label{tab:debias_results_appendix}
    \vspace{-0.2cm}
\end{table}

\subsection{Computational Efficiency}
We further show our approach is time and memory-efficient. Table~\ref{tab:resource} shows per epoch training time and GPU memory use of BERT, BERT+OC and BERT+SOC on the Gab corpus. We use one GeForce RTX 2080 Ti GPU to train each model. The training times of BERT+SOC and BERT+OC are only 4 times and 2 times of the original BERT. It is in contrast to the explanation regularization approach in~\cite{liu2019incorporating}, where it is reported to require 30x training time for the reported results on shallow CNN models. The inefficiency is introduced by the gradients over gradients, as also pointed out by~\cite{rieger2019interpretations}. Besides, our approach introduces only a small increase on the GPU memory use. 

\begin{table}[!ht]
\centering
\scalebox{0.85}{
\begin{tabular}{@{}lcc@{}}
\toprule
\textbf{Methods} & \textbf{Training time} & \textbf{GPU memory use} \\ \midrule
\textbf{BERT}               & 5 m 1 s                & 9095 M                  \\
\textbf{BERT+OC}            & 12 m 36 s              & 9411 M                  \\
\textbf{BERT+SOC}           & 19 m 38 s              & 9725 M                  \\ \bottomrule
\end{tabular}}
\caption{Per epoch training time of different methods on the Gab corpus. All methods finish training at around the third epoch.}
\label{tab:resource}
\end{table}

\end{document}